\documentclass[lettersize,journal]{IEEEtran}
\usepackage{cite}
\usepackage{footnote}
\usepackage{amsmath,bm}
\usepackage{amsfonts}
\usepackage{amssymb}
\usepackage{caption}
\usepackage{stfloats}
\usepackage{todonotes}
\usepackage{verbatim}
\usepackage{times}
\usepackage{epsfig}
\usepackage{graphicx}
\usepackage{subfigure}
\usepackage{overpic}
\usepackage{multirow}
\usepackage{mathtools}
\usepackage{algorithm}
\usepackage{algpseudocode}

\usepackage{setspace}
\usepackage{float}

\usepackage{lineno}
\begin{document}

% \linenumbers

\title{
Self-Error Adjustment: Theory and Practice of Balancing Individual Performance and Diversity in Ensemble Learning
    }

\author{
    Rui Zou
%    , Mengqi Wei, Shengyingjie Liu, Jianwen Sun, Sannyuya Liu
\IEEEcompsocitemizethanks{
\IEEEcompsocthanksitem

%This work was financially supported by the National Key R\&D Program of China (2022ZD0117103),
%National Natural Science Foundation of China (62293554),
%and Hubei Provincial Natural Science Foundation of China (2023AFA020).
%\textit{(Corresponding authors: Jianwen Sun, Sannyuya Liu)}

% R. Zou is with ... .
% Email: ... .
% All the other authors are with the 

%All authors are with the
%\{Faculty of Artificial Intelligence in Education,
%National Engineering Research Center of Educational Big Data\},
%Central China Normal University, Wuhan, China.
Gaoling School of Artificial Intelligence, Renmin University of China

Email: 
%\{zouruixyz, weimengqi, lsyj\}@mails.ccnu.edu.cn, \{sunjw, liusy027\}@ccnu.edu.cn.
zouruiwxyz@gmail.com
}
}

\markboth{Journal of \LaTeX\ Class Files,~Vol.~14, No.~8, August~2021}%
{Shell \MakeLowercase{\textit{et al.}}: A Sample Article Using IEEEtran.cls for IEEE Journals}


\maketitle

\begin{abstract}

Ensemble learning boosts performance by aggregating predictions from multiple base learners. A core challenge is balancing individual learner accuracy with diversity. Traditional methods like Bagging and Boosting promote diversity through randomness but lack precise control over the accuracy-diversity trade-off. Negative Correlation Learning (NCL) introduces a penalty to manage this trade-off but suffers from loose theoretical bounds and limited adjustment range. To overcome these limitations, we propose a novel framework called Self-Error Adjustment (SEA), which decomposes ensemble errors into two distinct components: individual performance terms, representing the self-error of each base learner, and diversity terms, reflecting interactions among learners. This decomposition allows us to introduce an adjustable parameter into the loss function, offering precise control over the contribution of each component, thus enabling finer regulation of ensemble performance. Compared to NCL and its variants, SEA provides a broader range of effective adjustments and more consistent changes in diversity. Furthermore, we establish tighter theoretical bounds for adjustable ensemble methods and validate them through empirical experiments. Experimental results on several public regression and classification datasets demonstrate that SEA consistently outperforms baseline methods across all tasks. Ablation studies confirm that SEA offers more flexible adjustment capabilities and superior performance in fine-tuning strategies.

\end{abstract}


\section{Introduction}
\label{sec:introduction}

\IEEEPARstart{E}{nsemble}
learning is a fundamental technique in machine learning, known for achieving superior performance by aggregating the predictions of multiple base learners \cite{zhou2012ensemble, dietterich2000ensemble}. This approach has demonstrated notable success across a wide range of fields, including image recognition \cite{krizhevsky2012imagenet}, speech recognition \cite{hinton2012deep}, natural language processing \cite{devlin2018bert}, healthcare \cite{mahajan2023ensemble}, environmental monitoring \cite{izonin2023cascade}, and large language models \cite{lai2024adaptive}.

One of the central challenges in constructing high-performance ensemble models is balancing the accuracy of individual learners with the diversity among them, which has been a core topic in ensemble learning theory \cite{kuncheva2003measures, brown2005diversity}. Diversity among base learners plays a critical role in enhancing ensemble performance \cite{hansen1990neural}, as increased diversity helps mitigate overfitting and improve the generalization capabilities of the ensemble model.
To further understand the role of diversity in ensemble learning, various error decomposition frameworks have been proposed. These frameworks break down ensemble error into components representing individual performance and diversity. For instance, \cite{krogh1994neural} decomposed generalization error into an error term and an ambiguity term, while \cite{ueda1996generalization} introduced the bias-variance-covariance decomposition. Building on these concepts, \cite{brown2005managing} connected different decomposition methods and enhanced the Negative Correlation Learning (NCL) approach. More recently, works such as \cite{steyvers2022bayesian} and \cite{zou2024balancing} utilized probabilistic and statistical theories to decompose ensemble error through multiplicative and additive rules, emphasizing the importance of diversity in human-machine collaboration.

Traditional ensemble methods like Bagging \cite{breiman1996bagging}, Boosting \cite{freund1996experiments}, and Random Forests \cite{breiman2001random} foster diversity indirectly by introducing randomness into the data or model space. However, these methods control diversity implicitly, making it challenging to achieve a precise balance between individual performance and diversity \cite{zhou2012ensemble}.
Negative Correlation Learning (NCL) \cite{liu1999ensemble} addresses this issue by introducing a penalty term into the loss function to explicitly regulate the trade-off between individual performance and diversity. While NCL has shown practical success, it has also led to several improved variants, such as Adaptive Negative Correlation Learning \cite{sheng2017niching}, Regularized Negative Correlation Learning \cite{chen2009regularized}, and Deep Negative Correlation Learning \cite{chen2010multiobjective, shi2018crowd}.

However, despite its success, NCL-based methods have certain limitations. First, their theoretical boundary conditions are relatively loose, leading to a large theoretical bound. Second, in practical applications, the range for effective adjustment is constrained, with substantial gaps between the achieved performance and the theoretical bound. Lastly, the adjustment strategy does not result in uniform changes in diversity, potentially causing the model to miss critical performance points.
Moreover, most existing ensemble error decomposition theories are derived from a statistical framework, which lacks an intuitive interpretation of the training process. 

To address these issues, we propose a novel framework called Self-Error Adjustment (SEA). SEA decomposes the ensemble error from the perspective of training, breaking it into two components: individual performance terms representing the self-error of each base learner, and diversity terms capturing the interactions among learners. Based on this decomposition, we introduce an adjustable parameter into the loss function, enabling precise control over the ratio between individual performance and diversity. This allows for more flexible adjustment of ensemble performance.

\begin{figure}[tb]
    \centering
    \includegraphics[width=0.42\textwidth]{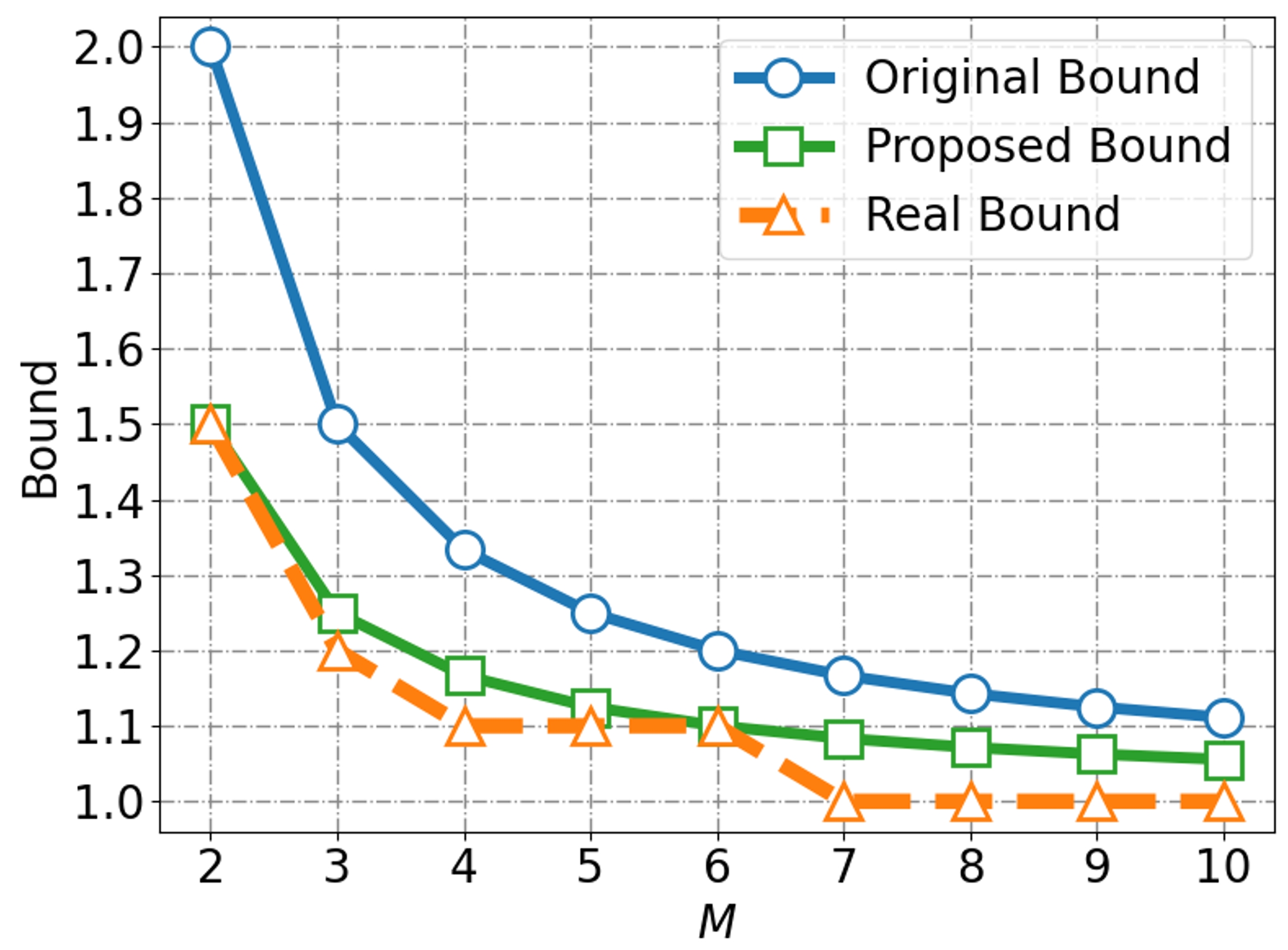}
    \caption{
        A comparison of the traditional theoretical bound (Original Bound), the proposed theoretical bound (Proposed Bound), and the actual bound (Real Bound). Adjustable parameters are incremented by 0.1, leading to an approximate estimate of the theoretical bound, which causes the actual bound to exhibit a step-like structure.
    }
    \label{fig:intro}
\end{figure}

Through theoretical analysis, we derive strict bounds for general adjustable ensemble methods under the SEA framework. In traditional settings, the boundary condition is based on the ``optimizability of the loss function", typically ensured by the positive definiteness of the Hessian matrix \cite{brown2005managing}. While this condition is usually equivalent to ``training can reduce ensemble error" for most tasks, it may not always hold in the context of adjustable ensemble learning. By considering both conditions, we establish a tighter theoretical bound than that of conventional methods. Figure~\ref{fig:intro} illustrates the relationship between the traditional theoretical bound (Original Bound), our proposed bound (Proposed Bound), and the actual bound (Real Bound) as the ensemble size $M$ increases. As the figure shows, the actual bound aligns more closely with our proposed bound than with the traditional theoretical bound.

In addition, we theoretically demonstrate that SEA provides a broader range of effective adjustments and more consistent diversity changes compared to NCL and its enhanced variant, NCL$^*$ \cite{brown2005managing}. This conclusion is further supported by our ablation experiments. In these experiments, we compare SEA with several mainstream ensemble methods across multiple public regression and classification datasets. The results consistently show that SEA achieves superior performance in all tasks.

Our main contributions are as follows:
(1) To the best of our knowledge, we are the first to decompose ensemble error from the training perspective and propose the SEA framework, which effectively balances individual performance and diversity.
(2) We conduct a comprehensive analysis of the SEA framework, deriving tighter theoretical bounds for adjustable ensemble methods and validating them experimentally. This leads to improved model training efficiency.
(3) We theoretically and empirically show that the advantages of SEA lie in its broader adjustment range and more uniform changes in diversity.
(4) We evaluate SEA on multiple regression and classification datasets, and the experimental results demonstrate that SEA consistently outperforms baseline methods in both task types.

The remainder of this paper is organized as follows:
Section \ref{sec:related_works} reviews the related work;
Section \ref{sec:method} details the SEA framework and provides the theoretical analysis;
Section \ref{sec:experiments} presents the experimental results and discussions;
Section \ref{sec:conclusion} concludes the paper and outlines future work.

\section{Related Work}\label{sec:related_works}

Ensemble learning, which improves generalization and robustness by combining multiple base learners, is a key area of research in machine learning. This section introduces classical ensemble learning methods, ensemble error decomposition theories, and Negative Correlation Learning (NCL) along with its variants.

\textbf{Classical Ensemble Learning Methods.} 
Bagging \cite{breiman1996bagging} creates multiple sub-datasets by bootstrapping the original training set. A base learner is independently trained on each sub-dataset, and their predictions are then averaged (for regression tasks) or combined via majority voting (for classification tasks). Bagging effectively reduces model variance and improves stability, making it particularly effective for high-variance learning algorithms such as decision trees. It is commonly used in models like Random Forests \cite{breiman2001random}, which further enhance performance and resistance to overfitting. 
Boosting, on the other hand, is a sequential ensemble technique where weak learners are trained iteratively, with each subsequent learner focusing on correcting the mistakes made by the previous one. This process progressively improves the model’s overall accuracy. Classical Boosting algorithms include AdaBoost \cite{freund1996experiments} and Gradient Boosted Decision Trees (GBDT) \cite{friedman2001greedy}. More recent Boosting-based models, such as XGBoost \cite{chen2016xgboost}, LightGBM \cite{ke2017lightgbm}, and CatBoost \cite{prokhorenkova2018catboost}, introduce techniques like parallelization, feature sampling, and optimized algorithms, significantly improving both training speed and prediction performance. These models have become widely adopted across various machine learning tasks in both academia and industry.
Stacking \cite{wolpert1992stacked} differs from Bagging and Boosting by combining different types of base learners and training a meta-learner to determine how best to fuse their outputs. Unlike simple averaging or voting, Stacking leverages the complementarity between learners to capture more complex relationships. Additionally, with the rise of deep learning, ensemble methods have been applied to deep neural networks, where ensembling multiple deep models improves both accuracy and stability. In computer vision, for instance, model ensembles have delivered impressive results in tasks like image classification \cite{krizhevsky2012imagenet} and object detection \cite{ren2016faster}.

\textbf{Ensemble Error Decomposition Theories.} 
Understanding the sources of error in ensemble models is essential for designing effective ensemble strategies. Ensemble error is typically decomposed into two components: the error of individual learners and a measure of their diversity. The bias-variance decomposition \cite{geman1992neural} separates the expected error into bias, variance, and noise. Ensemble learning aims to reduce variance while maintaining or reducing bias, thus enhancing generalization performance. 
The error-ambiguity decomposition \cite{krogh1994neural} expresses ensemble error as the sum of individual error and ambiguity, where ambiguity reflects the degree of disagreement between learners. Increasing ambiguity among learners can reduce overall ensemble error, leading to better generalization. Finally, the bias-variance-covariance decomposition \cite{ueda1996generalization} incorporates the covariance between learners, representing ensemble error as a combination of bias, variance, and covariance. Reducing covariance (i.e., increasing diversity) effectively lowers ensemble error and improves performance.

\textbf{Negative Correlation Learning (NCL) and its Improvements.} 
Negative Correlation Learning (NCL) \cite{liu1999ensemble} explicitly promotes diversity among learners by incorporating a negative correlation term into the loss function. During training, NCL minimizes the weighted sum of individual losses and the correlation between learners, thereby enhancing diversity while maintaining accuracy. This balance improves the generalization ability of the ensemble model. 
Adaptive Negative Correlation Learning (ANCL) \cite{sheng2017niching} extends this approach by dynamically adjusting the weight of the negative correlation term based on the learners' performance, thus better balancing individual accuracy and diversity during training. 
Regularized Negative Correlation Learning \cite{chen2009regularized} introduces a regularization term to prevent overfitting while preserving diversity among learners. Multi-objective optimization strategies have also been employed to jointly optimize individual losses and the negative correlation term. Deep Negative Correlation Learning \cite{shi2018crowd} applies the NCL approach to deep neural networks, encouraging diversity among models during training and enhancing the performance of ensemble deep models. This method has shown success in tasks across domains such as computer vision and natural language processing. Additionally, Multi-task Negative Correlation Learning \cite{shi2018crowd} extends NCL to multi-task learning, where negative correlation terms are introduced across related tasks to promote information sharing and diversity, thereby improving overall learning performance.

\section{Method}\label{sec:method}

In this section, we first introduce the SEA framework. We then explore the theoretical bounds and applications of adjustable ensemble methods. Lastly, we investigate the key factors influencing the performance of adjustable ensemble methods and their effective adjustment ranges.

\subsection{Self-Error Adjustment Framework}\label{sec:SEA}

Consider a regression dataset
$\mathcal{L}=\{(t^{(n)}, \bm{x}^{(n)})\ |\ n=1,2,\ldots, N\}$, 
comprising $N$ samples, where 
$t^{(n)}\in \mathbb{R}$ represents the target value for the $n$-th sample. 
We have $M$ learners $\{f_i\ | i=1,\ldots,M\}$, where the prediction of the $i$-th learner is denoted as $f_i(\bm{x};\bm{\theta}_i)$, 
with $\bm{x}$ representing the input and $\bm{\theta}_i$ representing the trainable parameters of the $i$-th learner.

For convenience, we denote the prediction $f_i(\bm{x};\bm{\theta}_i)$ as $f_i$. 
The ensemble error $(\bar{f} - t)^2$ serves as the loss function, where:
    \begin{align}
        \label{eq:disagreementDecomposition_1} 
        e_i = (\bar{f} - t)^2,
    \end{align}  
with $\bar{f} = \frac{1}{M}\sum_{i=1}^{M}f_i$ representing the ensemble prediction. Since the trainable parameters $\bm{\theta}_i$ are encapsulated in $f_i(\bm{x};\bm{\theta}_i)$ or $f_i$, we can decompose Eq~\ref{eq:disagreementDecomposition_1} into components related to $f_i$:
\begin{align}
        \nonumber
        e_i=& 
        \frac{1}{M^2} 
        [\sum_{j\neq i} (f_j-t) + (f_i-t)]^2 
        \\ \nonumber  
        =&\frac{1}{M^2} 
        \{(f_i-t)^2 + 
        2(f_i-t)\sum_{j\neq i} (f_j-t)+
        \\ \label{eq:disagreementDecomposition_2}  
        &[\sum_{j\neq i} (f_j-t)]^2 \}.
    \end{align}  
Since the predictions of individual learners $f_i$ and $f_j$ are independent, i.e., $f_i$ and $f_j$ are independent for $i \neq j$, it follows that:
$
\frac{\partial e_i}{\partial f_j} = 0 \quad (\forall j \neq i).
$
Due to this independence, the term 
$
\left[\sum_{j \neq i} (f_j - t)\right]^2
$
does not affect the training process. Therefore, we can remove this constant term. Additionally, the multiplicative constant $\frac{1}{M^2}$ has no impact on training, so it can also be omitted. This simplifies the error to:
    \begin{align}
        \label{eq:111}
        e_i &= 
        (f_i - t)^2 + 
        2(f_i - t)\sum_{j \neq i} (f_j - t).
    \end{align}  
To further simplify Eq~\ref{eq:111}, we introduce the concept of ``complementary prediction". For the $i$-th learner, we define its complementary prediction as follows: when $f_i$ is replaced by its complementary prediction $g_i$, the ensemble error becomes zero, meaning the ensemble prediction after replacement exactly matches the target value $t$:
\begin{equation}
    \label{eq:f_complementary_1}
    \frac{1}{M}\left(\sum_{j \neq i} f_j + g_i\right) = t. 
\end{equation}
From this, the complementary prediction can be derived as:
\begin{align}
    \label{eq:f_complementary}
    g_i = M(t - \bar{f}) + f_i.
\end{align}
It is straightforward to observe that 
$
    \sum_{j \neq i}(f_j - t) = -(g_i - t),
$
and substituting this into Eq~\ref{eq:111} gives:
\begin{align}
    \label{eq:disagreementDecomposition_compliment_3}
    e_i = 
    (f_i - t)^2 - 2(f_i - t)(g_i - t). 
\end{align}

\textbf{Introducing Adjustable Parameter.}
In Eq~\ref{eq:111}, we refer to $(f_i - t)$ as the self-error term, representing the portion of the ensemble error attributable to individual learners. The performance of individual learners is directly related to this term: the smaller $(f_i - t)^2$ is, the smaller the individual error, indicating better performance. If we focus solely on the individual performance term, it is evident that improving individual performance leads to a reduction in ensemble error.

Diversity is related to the second part of the equation, specifically 
$
2(f_i - t)\sum_{j \neq i} (f_j - t),
$
where smaller values indicate greater diversity. For example, when the self-error term $(f_i - t)$ and $\sum_{j \neq i} (f_j - t)$ have opposite signs, it suggests that the current learner’s prediction differs from the others. The smaller this term becomes, the greater the difference (i.e., diversity) between the current learner and the ensemble. If we focus solely on the diversity term, it becomes clear that increased diversity reduces the ensemble error.

However, individual performance and diversity are not independent. To minimize the ensemble error, it is crucial to balance the relationship between these two components.
Therefore, in Eq~\ref{eq:disagreementDecomposition_compliment_3}, we introduce an adjustable parameter $k$ to control the trade-off between the term representing individual performance $(f_i - t)^2$ and the term representing diversity $(f_i - t)(g_i - t)$. This results in an adjustable loss function for SEA, denoted as $e_i^{SEA}$:
\begin{equation}
    \label{eq:e_div_disagreementDecomposition_comp} 
    e_i^{SEA} =
    (f_i - t)^2 - 
    2k(f_i - t)(g_i - t). 
\end{equation}
For better understanding and clarity, we reformulate this into a vector form. We define the vector $\overrightarrow{tf_i} = f_i - t$, and similarly, $\overrightarrow{tg_i} = g_i - t$. A constant term $[k(g_i - t)]^2$, which does not affect the training process, is added to complete the square. This leads to the following vector form:
\begin{equation}
    \label{eq:e_div_disagreementDecomposition_comp_vec}
    e_i^{SEA} =
    (\overrightarrow{tf_i} - k*\overrightarrow{tg_i})^2.
\end{equation}
Eq~\ref{eq:e_div_disagreementDecomposition_comp_vec} illustrates that the training objective of SEA is to adjust the relative positions of $\overrightarrow{tf_i}$ and $\overrightarrow{tg_i}$. Using the training target $t$ as the reference point, the behavior of $f_i$ and $g_i$ depends on the value of $k$. 

When \textit{$k > 0$}, the training objective is for $f_i$ and $g_i$ to move in the same direction from $t$. In particular, when \textit{$k = 1$}, $f_i$ and $g_i$ converge, minimizing the ensemble error as defined by the complementary prediction $g_i$. When \textit{$k = 0$}, the focus shifts entirely to minimizing the individual error, causing $f_i$ to move directly towards $t$. Conversely, when \textit{$k < 0$}, $f_i$ and $g_i$ are encouraged to move in opposite directions from $t$, which, according to the complementary prediction definition, leads to an increase in ensemble error.
A more detailed theoretical proof of these behaviors is provided in the boundary analysis (Section~\ref{sec:boundary}). Algorithm~\ref{alg:SEA} outlines the implementation of the SEA algorithm.

\begin{algorithm}[tb]
\caption{SEA Algorithm Implementation}
\label{alg:SEA}
\begin{spacing}{1.3} % 1.5x line spacing
\begin{algorithmic}[1]
    \Require Dataset $\mathcal{L} = \{(t^{(n)}, \bm{x}^{(n)})\ |\ n=1,2,\ldots,N\}$
    \State Initialize $M$ base learners $\{f_i \ |\ i=1,\ldots,M\}$
    \State Compute the average prediction of the base learners $\bar{f}=\frac{1}{M}\sum_{i=1}^{M} f_i$
    \For{$i = 1$ to $M$}
        \State Compute the complementary prediction $g_i = M(t - \bar{f}) + f_i$
        \State Compute the loss function $e_i^{SEA} = \frac{1}{2} (\overrightarrow{tf_i} - k * \overrightarrow{tg_i})^2$
    \EndFor
    \State Aggregate the total loss $e^{SEA} = \sum_{i=1}^{M} e_i^{SEA}$
    \For{$i = 1$ to $M$}
        \State Update $f_i(\bm{x}; \theta_i)$ by $\theta_i := \theta_i - \alpha \frac{\partial e^{SEA}}{\partial \theta_i}$
    \EndFor
    \State \textbf{Return} the final prediction $\bar{f}=\frac{1}{M}\sum_{i=1}^{M} f_i$
\end{algorithmic}
\end{spacing}
\end{algorithm}

\subsection{Boundary Analysis of Adjustable Ensemble Methods}
\label{sec:boundary}

We now examine the theoretical boundaries of adjustable ensemble methods within the SEA framework.

In traditional adjustable ensemble methods, boundary conditions are typically assumed to be similar to those in deep learning, where the loss function is optimizable \cite{brown2003use,brown2005managing}. While this assumption holds for most deep learning tasks, adjustable ensemble learning presents more complex boundary conditions. This complexity arises because optimizing the loss function can, at times, lead to an increase in ensemble error.
Thus, we redefine the boundary conditions for adjustable ensemble methods as follows:
(1) The loss function must be optimizable;
(2) The ensemble error must decrease after training.

Let us examine these two boundary conditions within the SEA framework. Boundary condition (1) is inherently satisfied because the Hessian matrix of Eq~\ref{eq:e_div_disagreementDecomposition_comp} with respect to $f_i$ is always positive definite. Specifically, the Hessian matrix $H=2 > 0$, ensuring that the loss function has a local minimum \cite{goodfellow2016deep}. This confirms that the loss function in Eq~\ref{eq:e_div_disagreementDecomposition_comp} is optimizable under this condition.

Next, we consider boundary condition (2). We express $f_i$ and $g_i$ in terms of $\bar{f}$ as follows:
$$
    f_i = M\bar{f} - \sum_{j \neq i}f_j, \quad
    g_i = M(t - \bar{f}) + f_i. 
$$
Substituting these expressions into the SEA loss function in Eq~\ref{eq:e_div_disagreementDecomposition_comp}, we obtain:
\begin{align}
    \label{eq:e_div_disagreementDecomposition_comp_vec_aver}
    e_i^{SEA} = 
    \left[M*\overrightarrow{t\bar{f}} - (1-k)(M-1) \frac{1}{M-1}\sum_{j \neq i}\overrightarrow{tf_j}\right]^2.
\end{align}
This equation shows that the optimization objective of SEA is:
\begin{align}
    \nonumber
    \overrightarrow{t\bar{f}} 
    &\to
    \frac{(1-k)(M-1)}{M} *  
    \frac{1}{M-1}\sum_{j \neq i}\overrightarrow{tf_j}. 
\end{align}
Here, $\to$ indicates that the optimization goal is for the former expression to approach the latter. For simplicity, we define $\beta = \frac{(1-k)(M-1)}{M}$. The equation then simplifies to:
\begin{align}
    \label{eq:target_2}
    \overrightarrow{t\bar{f}} 
    &\to
    \beta * 
    \frac{1}{M-1}\sum_{j \neq i}\overrightarrow{tf_j}. 
\end{align}
Here, 
$\overrightarrow{t\bar{f}}=\frac{1}{M}\sum_{i=1}^{M}\overrightarrow{tf_i}$ 
represents the average error, while $\frac{1}{M-1}\sum_{j \neq i}\overrightarrow{tf_j}$ 
is the average error excluding $f_i$.
In general, these two quantities can be considered approximately equal:
\begin{align}
    \label{eq:target_approx}
    \frac{1}{M-1}\sum_{j \neq i} \overrightarrow{tf_j} \approx 
    \frac{1}{M}\sum_{j=1}^{M} \overrightarrow{tf_j} = 
    \overrightarrow{t\bar{f}}. 
\end{align}
This approximation is more accurate when $M$ is large or when $\left | k \right |$ is small. As $M$ increases, the contribution of $\overrightarrow{tf_i}$ to $\overrightarrow{t\bar{f}}$ diminishes, making the approximation more valid. Similarly, as $\left | k \right |$ decreases, the differences between individual learners also diminish. For instance, when $k=0$, the individual training objectives are identical, minimizing differences between learners and further validating the approximation.

The newly introduced boundary condition—that the ensemble error must always decrease after training—implies that the target ensemble error ($\left | \beta * \frac{1}{M-1}\sum_{j \neq i}\overrightarrow{tf_j} \right |$) should be smaller than the current ensemble error ($\left | \overrightarrow{t\bar{f}} \right |$). This gives the following inequality:
\begin{align}
    \label{eq:target_better_1}
    \left | \beta * \frac{1}{M-1}\sum_{j \neq i}\overrightarrow{tf_j} \right |
    <
    \left | \overrightarrow{t\bar{f}} \right |. 
\end{align}
Substituting Eq~\ref{eq:target_approx} into the left-hand side of the inequality, we derive the theoretical boundary condition:
$$
    \left | \beta \right | < 1.
$$
From $\beta = \frac{(1-k)(M-1)}{M}$, we can then derive the theoretical boundary for the SEA adjustment parameter $k$:
\begin{align}
    \label{eq:target_better_3}
    \frac{-1}{M-1} < k < 2 + \frac{1}{M-1}. 
\end{align}

Compared to the boundary derived by considering only boundary condition (1) 
($-\infty < k < +\infty$), 
we obtain a tighter boundary 
($\frac{-1}{M-1} < k < 2 + \frac{1}{M-1}$) 
by incorporating both boundary conditions (1) and (2).

\subsection{Theoretical Bound of Adjustable Ensemble Methods}
\label{sec:boundary_old}

For any adjustable ensemble method, the boundaries can be calculated using the two boundary conditions outlined above. Here, we provide a simple algorithm based on the SEA framework to demonstrate this process. The algorithm involves analyzing the relationship between the adjustable ensemble method and SEA, and then deriving the boundaries for other methods based on SEA’s boundary. We illustrate this approach with classical adjustable ensemble methods such as NCL and NCL$^*$.
As described in the related work, NCL \cite{liu1997negatively,liu1999ensemble,liu1999simultaneous} explicitly adjusts the balance between individual accuracy and diversity by modifying the penalty coefficient, making it one of the most widely used adjustable ensemble methods. NCL$^*$ \cite{brown2003use,brown2005managing} corrected an error in the original NCL assumptions and established a connection between NCL$^*$ and the error-ambiguity decomposition, offering a method for calculating the boundaries of general adjustable ensemble methods.

\textbf{NCL$^*$}. 
We first derive the boundary for NCL$^*$, whose loss function for the $i$-th learner is given by:
\begin{equation}
    \label{eq:NCL*_e_i_div}
    e_i^{NCL^*} = \frac{1}{2}(f_i - t)^2 - \gamma \frac{1}{2}(f_i - \bar{f})^2. 
\end{equation}
Here, $\gamma$ is the adjustment parameter that controls the weight of the penalty term $\frac{1}{2}(f_i - \bar{f})^2$, which encourages diversity. The ensemble prediction is the average of the individual predictions, 
$\bar{f} = \frac{1}{M}\sum_{i=1}^{M}f_i$.

First, we convert the above loss function into a polynomial form similar to Eq~\ref{eq:e_div_disagreementDecomposition_comp}:
\begin{align}
\nonumber
    e_i^{NCL^*} =&
    \frac{1}{2}\left[1 - \gamma \left(1 - \frac{1}{M}\right)^2\right]
    (f_i - t)^2 -
    \\  \label{eq:NCL_e_i_div_SEA}
    &\frac{\gamma}{M}\left(1 - \frac{1}{M}\right)
    (g_i - t)(f_i - t) + C_1, 
\end{align}
where the constant term $C_1$ does not involve $f_i$, and $C_1 = -\frac{\gamma}{2M^2}(g_i - t)^2$.
The adjustable parameter $\gamma$ controls the ratio between the first and second terms. By comparing the ratio of the coefficients in the above equation with those in Eq~\ref{eq:e_div_disagreementDecomposition_comp}, we can determine the boundary condition.
The relationship between the adjustable parameter $\gamma$ in NCL$^*$ and the adjustable parameter $k$ in SEA is given by the following ratio:
\begin{align}
    \label{eq:k_NCL_1}
    \frac{-2k}{1} =
    \frac
        {-\frac{\gamma}{M}\left(1-\frac{1}{M}\right)}
        {\frac{1}{2}\left[1-\gamma \left(1-\frac{1}{M}\right)^2\right]}. 
\end{align}
Solving for $\gamma$, we get:
\begin{equation}
    \label{eq:k_NCL}
    \gamma = \frac{kM^2}{(M-1)\left[k(M-1) + 1\right]}. 
\end{equation}
Based on the SEA boundary 
($\frac{-1}{M-1} < k < 2 + \frac{1}{M-1}$), 
we can derive the theoretical boundary for NCL$^*$, denoted as $\gamma_{SEA}$ for the corrected theoretical boundary:
\begin{equation}
    \label{eq:NCL_2_gamma}
    \gamma_{SEA} < \frac{M \left(M - \frac{1}{2}\right)}{\left(M-1\right)^2}. 
\end{equation}
The boundary provided by condition (1) alone is:
\begin{equation}
    \label{eq:NCL_2_gamma_}
    \gamma_1 < \left(\frac{M}{M-1}\right)^2. 
\end{equation}
It is important to note that 
$
\gamma_{SEA} < \gamma_1,
$
which indicates that the boundary derived from the SEA framework is tighter than the one obtained by considering only condition (1).

\textbf{NCL}.
For adjustable ensemble methods with more complex loss functions, the computation can be handled through differentiation and integration. Taking NCL \cite{brown2003use} as an example, we demonstrate how to manage more complex loss functions. The loss function for NCL is given by:
\begin{equation}
    \label{eq:NCLorigina_e_i}
    e_i^{NCL} = \frac{1}{2} (f_i - t)^2 + \lambda p_i, 
\end{equation}
where $\lambda$ is the adjustable parameter that controls the balance between the penalty term representing diversity $p_i = (f_i - \bar{f}) \sum_{j \neq i}(f_j - \bar{f})$ and the individual performance term $(f_i - t)^2$. Since directly converting the equation into a polynomial form is complex, we first take the derivative:
\begin{align}
    \frac{\partial e_i^{NCL}}{\partial f_i} =
    (f_i - t) - \lambda (f_i - \bar{f}).
\end{align}
By integrating the above equation, we obtain:
\begin{align}
    e_i^{NCL} =&
    \frac{1}{2}\left[1 - \lambda \left(1 - \frac{1}{M}\right)\right] 
    (f_i - t)^2 -
    \\ \nonumber
    &\frac{\lambda}{M}
    (g_i - t)(f_i - t) + C_2, 
\end{align}
where $C_2$ is the integration constant.
As in previous cases, we compare the ratio of the coefficients in the above equation with those in Eq~\ref{eq:e_div_disagreementDecomposition_comp}:
\begin{align}
    \frac{-2k}{1} =
    \frac
        {-\frac{\lambda}{M}}
        {\frac{1}{2}\left[1 - \lambda \left(1 - \frac{1}{M}\right)\right]}. 
\end{align}
The relationship between the adjustable parameter $\gamma$ in NCL$^*$ and the adjustable parameter $k$ in SEA is given by:
\begin{align}
    \label{eq:k_NCLorginal}
    \lambda = \frac{kM}{1+k(M-1)}. 
\end{align}
From the boundary $\frac{-1}{M-1} < k < 2 + \frac{1}{M-1}$, we can derive the NCL boundary, denoted as:
\begin{align}
    \lambda_{SEA} < \frac{2M - 1}{2(M - 1)}.
\end{align}
On the other hand, when considering only boundary condition (1), which is derived from the positive definiteness of the Hessian matrix, the NCL boundary is given by:
\begin{equation}
    \label{eq:NCL_2_lambda}
    \lambda_1 < \frac{M}{M-1}. 
\end{equation}
Notably, 
$
\lambda_{SEA} < \lambda_1,
$
indicating that the boundary derived from the SEA framework is tighter.

\begin{figure}[tb]
    \centering
    \includegraphics[width=0.486\textwidth]{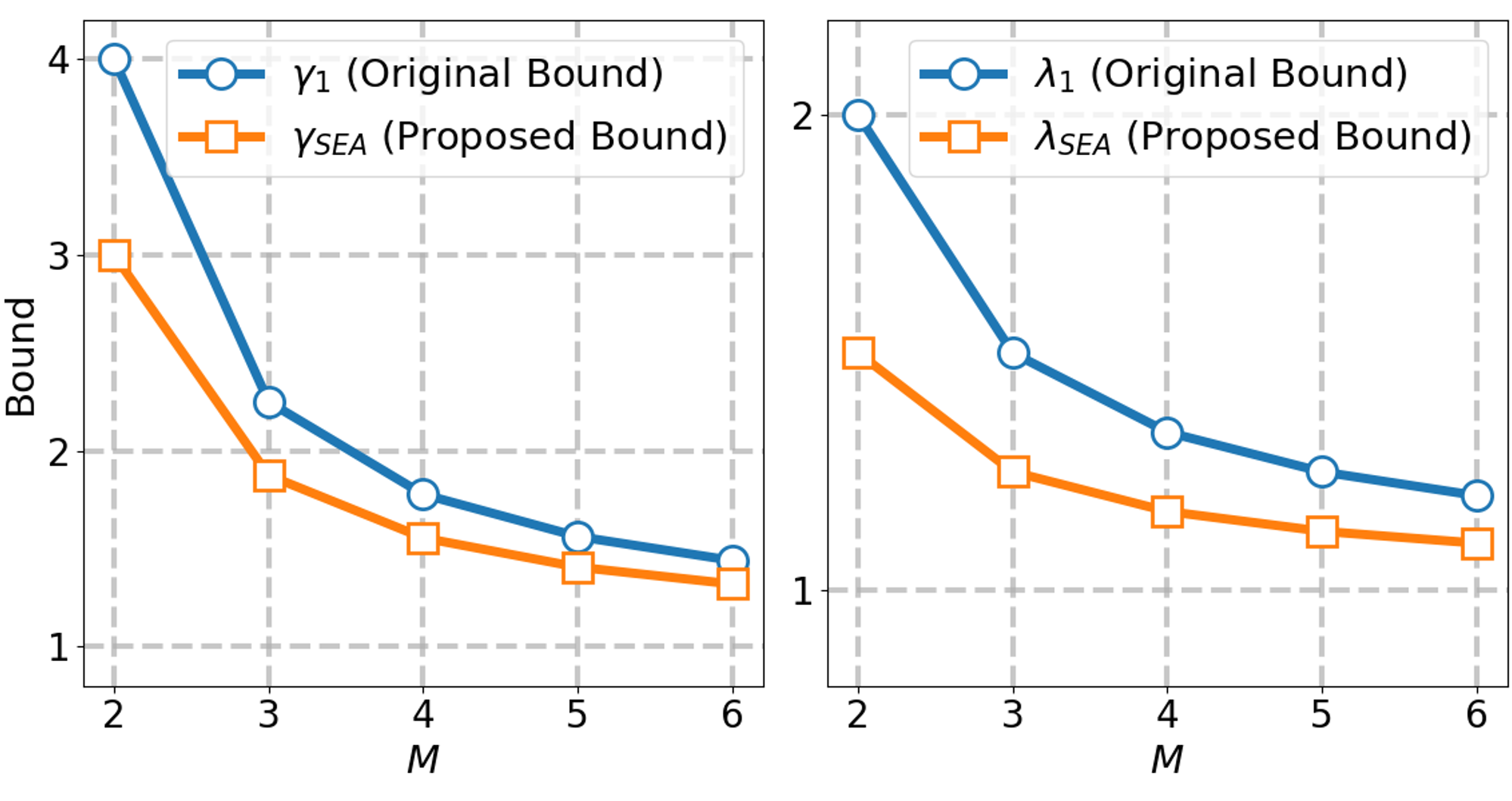}
    \caption{
        Comparison of theoretical boundaries. 
        $\gamma_1$ and $\lambda_1$ represent the theoretical boundaries calculated by considering only boundary condition (1) and the Hessian matrix;
        $\gamma_{SEA}$ and $\lambda_{SEA}$ represent the theoretical boundaries calculated by considering both boundary conditions (1) and (2) based on the SEA framework.
    }
    \label{fig:theoryBoundary}
\end{figure}

From the above calculations, we observe that traditional research, which considers only boundary condition (1) and derives the boundary from the positive definiteness of the Hessian matrix, does not yield the tightest boundary. This discrepancy becomes more pronounced as the ensemble size $M$ decreases, as illustrated in Figure~\ref{fig:theoryBoundary}.

\subsection{Study on the Effective Adjustment Range}

Although the theoretical boundaries have been established, they do not directly translate into the effective adjustment range used in practical applications. For example, while the theoretical boundary for NCL is 
$
\lambda \in (-\infty, \frac{2M - 1}{2(M - 1)}),
$
in reality, setting $\lambda$ to $-\infty$ is impractical. Furthermore, 
$
\frac{2M - 1}{2(M - 1)}
$
is not a constant value, so researchers typically approximate this boundary with a constant value close to it. According to previous studies \cite{brown2003use,brown2005managing}, the effective adjustment range for NCL and NCL$^*$ is generally set to $[0, 1]$. For example, in \cite{liu1997negatively}, $\lambda$ is selected from the set $\{0, 0.25, 0.5, 1\}$.

However, a limitation of this approach is that the effective adjustment range is much smaller than the full theoretical range.
In contrast, for SEA, this issue has a smaller impact. The theoretical boundary for SEA is $k \in \left(\frac{-1}{M-1}, 2 + \frac{1}{M-1}\right)$, and this boundary can be approximated by selecting constants close to it, setting the effective adjustment range as $R_{SEA} = [0, 2]$.

We can also convert the effective adjustment range of NCL and NCL$^*$ into the SEA framework for comparison. For instance, based on the relationship between the adjustable parameters $\gamma$ in NCL$^*$ and $k$ (Eq~\ref{eq:k_NCL}), we derive:
\begin{align}
    \label{eq:k_NCL_2}
    k =
    \frac{(M-1)\gamma}{M^2-(M-1)^2\gamma}.
\end{align}
By substituting the effective adjustment range of NCL$^*$, $[0, 1]$, into the above equation, we can obtain its effective adjustment range in the SEA framework as:
\[
R_{NCL^*} \in \left[0, \frac{M-1}{2M-1}\right].
\]

Similarly, from the relationship between the adjustable parameter $\lambda$ in NCL and $k$ (Eq~\ref{eq:k_NCLorginal}), we derive:
\begin{align}
    \label{eq:lambda_k_NCLorginal}
    k = \frac{\lambda}{M - \lambda (M-1)},
\end{align}
By substituting the effective adjustment range of NCL, $[0, 1]$, into the above equation, we can obtain its effective adjustment range in the SEA framework as:
\[
R_{NCL} \in [0, 1].
\]

\begin{figure}[tb]
    \centering
    \includegraphics[width=0.42\textwidth]{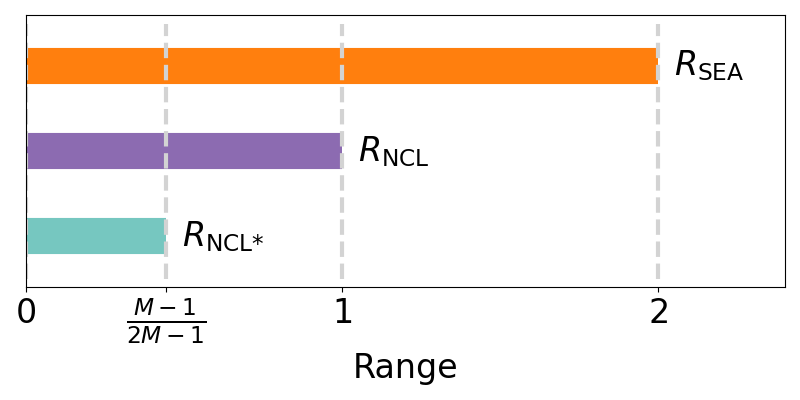}
    \caption{
        Comparison of effective adjustment ranges. 
        We compare the effective adjustment ranges of SEA, NCL, and NCL$^*$ within the SEA framework.
    }
    \label{fig:range}
\end{figure}

Figure~\ref{fig:range} compares the effective adjustment ranges of the three methods within the SEA framework. It can be observed that their relationship is:  
\(
R_{SEA} \supset R_{NCL} \supset R_{NCL^*}.
\)   
This, to some extent, determines the performance ranking of the three methods:
\[
SEA \succ NCL \succ NCL^*.
\]
$\succ$ indicates that the former method outperforms the latter, a conclusion further supported by our experimental results.

\subsection{Study on Diversity Variation}
\label{app:theorem}

In addition to the advantages observed within the broader adjustment range, as shown in the ablation study (see Section~\ref{sec:advantage}), we further found that SEA continues to outperform baseline models even when restricted to the same effective adjustment range. We hypothesize that this superior performance is closely tied to the characteristics of diversity variation: SEA's adjustment strategy facilitates more "uniform" diversity changes in response to adjustable parameters compared to baseline methods. This uniformity allows the ensemble method to more effectively explore a wide range of performance and diversity combinations across individual learners, thus increasing the likelihood of discovering an optimal ensemble configuration.

In regression tasks, diversity is often quantified by the standard deviation (\(\text{std}\)) of the predictions among individual learners. To further investigate this, we examined how the \(\text{std}\) evolves under the SEA framework as the adjustable parameters change. We also extended our analysis to evaluate the diversity performance of general adjustable ensemble learning methods.

For the SEA ensemble framework, let there be $M$ individual models, where the prediction of the $i$-th learner is denoted by $f_i\ (i=1,\ldots, M)$. The standard deviation of the $M$ predictions is defined as follows:
\begin{equation}
\label{eq:std}
\text{std} = \sqrt{
\frac{1}{M} \sum_{i=1}^{M} (f_i - \bar{f})^2
}.
\end{equation}
Here, $\bar{f}=\frac{1}{M}\sum_{i=1}^{M}f_i$ represents the mean prediction of all learners. Following the approach used in Eq~\ref{eq:e_div_disagreementDecomposition_comp_vec}, we can express $(f_i - \bar{f})$ in vector form as:
\begin{equation}
\label{eq:k_std_2}
f_i - \bar{f} = 
\overrightarrow{tf_i} - \overrightarrow{t\bar{f}} 
= 
(M-1) \left(
\frac{1}{M} \sum_{i=1}^{M} \overrightarrow{tf_i} - 
\frac{1}{M-1} \sum_{i \neq j} \overrightarrow{tf_j}
\right).
\end{equation}
This equation provides a vector-based decomposition of the standard deviation, which facilitates a deeper understanding of the diversity variation within the SEA framework.
The training objective of SEA (see Eq.~\ref{eq:e_div_disagreementDecomposition_comp_vec_aver})
is substituted into Eq.~\ref{eq:k_std_2}, yielding:
\begin{equation}
\label{eq:k_std_3}
f_i - \bar{f} = -\frac{M-1}{M} \left[1 + (M-1)k\right]
\frac{1}{M-1} \sum_{j \neq i} \overrightarrow{tf_j}.
\end{equation}
Next, by substituting this result into the expression for the standard deviation in Eq.~\ref{eq:std}, we obtain:
\begin{equation}
\label{eq:k_std_4}
\text{std} = \frac{1}{\sqrt{M}} \frac{M-1}{M} \left[1 + (M-1)k\right]
\sqrt{\sum_{i=1}^{M} \left(
\frac{1}{M-1} \sum_{j \neq i} \overrightarrow{tf_j}
\right)^2}.
\end{equation}

Using the approximation condition in Eq.~\ref{eq:target_approx}, and substituting it into the above formula, we can simplify it further as:
\begin{equation}
\label{eq:k_std_5}
\text{std} = \frac{M-1}{M} [1 + (M-1)k] \left|\overrightarrow{t\bar{f}}\right|.
\end{equation}
Within the effective adjustment range, the ensemble error can be significantly reduced, leading to minimal differences in ensemble error for varying adjustable parameters. We can assume that these ensemble errors are approximately equal, i.e.,
\[
\left|\overrightarrow{t\bar{f}}|_{k=k_m}\right| \approx \left|\overrightarrow{t\bar{f}}|_{k=k_n}\right| = C,
\]
where $k_m$ and $k_n$ are any two parameters within the effective adjustment range, and $C$ is a constant. Substituting this approximation into Eq.~\ref{eq:k_std_4}, we obtain:
\begin{equation}
\label{eq:k_std_6}
\text{std}(k) = \frac{M-1}{M} [1 + (M-1)k] C,
\end{equation}
where $\frac{M-1}{M} [1 + (M-1)k]$ is a linear function of $k$, implying that the standard deviation $\text{std}(k)$ exhibits a uniform linear trend with respect to changes in $k$.

\textbf{NCL.}  
We can apply the SEA framework to study the diversity variation in general adjustable ensemble methods. Taking NCL as an example, Eq.~\ref{eq:k_NCLorginal} defines the relationship between $\lambda$ and $k$:
\[
\lambda = \frac{kM}{1+k(M-1)}.
\]
By substituting Eq.~\ref{eq:k_NCLorginal} into Eq.~\ref{eq:k_std_6}, we derive the relationship between diversity (standard deviation $\text{std}$) and the adjustable parameter $\lambda$ in NCL:
\begin{equation}
\label{eq:std_lambda_relation}
\text{std}(\lambda) = -\frac{M-1}{\lambda(M-1) - M} C.
\end{equation}
This expression reveals that $\text{std}(\lambda)$ is nonlinear. To further characterize this nonlinearity, we calculate its curvature:
\[
R=\left| \text{std}''(\lambda) \right| = \frac{2C}{\left| (\lambda - 1) - \dfrac{1}{M-1} \right|^3}.
\]
Within the effective adjustment range ($0 \leq \lambda \leq 1$), as $M$ increases, the curvature $R$ also increases, indicating greater nonlinearity. This result suggests that as the ensemble size ($M$) grows, the diversity in NCL (measured by the standard deviation $\text{std}$) becomes increasingly uneven with respect to changes in the adjustable parameter ($\lambda$). This finding aligns with our experimental results.

\begin{table}[tb]
    \centering
\begin{tabular}{lccl}
\hline
Dataset    & Samples & Features & Source \\ \hline
pyrim               & 74    & 27   & UCI\cite{pyrim_UCI}    \\
triazines           & 186   & 60   & UCI\cite{pyrim_UCI} \\
bodyfat             & 252   & 14   & StatLib\cite{bodyfat_StatLib} \\
eunite2001          & 367   & 16   & Eunite 2001\cite{chen2004load} \\
mpg                 & 392   & 7    & UCI\cite{pyrim_UCI} \\
housing             & 506   & 13   & UCI\cite{pyrim_UCI} \\
mg                  & 1,385 & 6    & GWF01a\cite{flake2002efficient} \\
abalone             & 4,177 & 8    & UCI\cite{pyrim_UCI} \\
cpusmall            & 8,192 & 12   & Delve\cite{cpusmall_Delve} \\ \hline
\end{tabular}
    \caption{
Summary statistics of the regression datasets. “Samples” refers to the total number of instances, and “Features” refers to the number of attributes in each dataset.
    }
    \label{tab:regression_datasets}
\end{table}

\begin{table}[tb]
    \centering
    \setlength{\tabcolsep}{4pt} % Adjust column spacing
    \begin{tabular}{lcccl}
\hline
Dataset   & Samples & Features & Classes & Source \\ \hline
sonar          & 208    & 60    & 2     & UCI\cite{gorman1988analysis} \\
ionosphere     & 351    & 34    & 2     & UCI\cite{sigillito1989classification} \\
vehicle        & 846    & 18    & 4     & StatLog\cite{michie1995machine} \\
german         & 1,000  & 24    & 2     & StatLog\cite{german_StatLog} \\
dna            & 3,186  & 180   & 3     & StatLog\cite{michie1995machine} \\
satimage       & 6,435  & 36    & 6     & StatLog\cite{satimage_StatLog} \\ \hline
\end{tabular}
    \caption{
Summary statistics of the classification datasets. “Classes” refers to the number of target categories in each dataset.
    }
    \label{tab:classification_datasets}
\end{table}

\begin{table*}[b]
\centering
\begin{tabular}{l|ccccccccc}
\hline
Datasets & Bagging & SSE & sGBM & NCL & NCL* & SEA & Improvement \\
\hline
pyrim & \underline{0.577} & 0.599 & 0.591 & 0.587 & 0.588 & \textbf{0.523} & 10.33\% \\
triazines & 0.931 & \underline{0.848} & 0.875 & 0.855 & 0.855 & \textbf{0.766} & 10.70\% \\
bodyfat & 0.19 & 0.308 & \underline{0.172} & 0.172 & 0.173 & \textbf{0.158} & 8.86\% \\
eunite2001 & 0.417 & 0.485 & 0.424 & \underline{0.413} & 0.415 & \textbf{0.382} & 8.12\% \\
mpg & 0.922 & 0.838 & 0.621 & \underline{0.605} & 0.66 & \textbf{0.503} & 20.28\% \\
housing & 0.447 & 0.51 & 0.443 & \underline{0.41} & 0.412 & \textbf{0.368} & 11.41\% \\
mg & 0.552 & 0.597 & 0.536 & \underline{0.525} & 0.532 & \textbf{0.481} & 9.15\% \\
abalone & 0.637 & 0.674 & 0.643 & 0.633 & \underline{0.633} & \textbf{0.601} & 5.32\% \\
cpusmall & 0.434 & 0.444 & 0.407 & \underline{0.394} & 0.405 & \textbf{0.334} & 17.96\% \\
\hline
\end{tabular}
\caption{
    RMSE results for regression tasks with ensemble sizes \(M \in \{5, 10, 20\}\) and 5-fold cross-validation. 
    Bold values indicate the best RMSE for each dataset.
    Underlined values represent the second-best RMSE.
    ``Improvement" refers to the percentage improvement of the best RMSE over the second-best RMSE. 
    For example, if the second-best RMSE is 0.588 and the best RMSE is 0.523, the relative improvement is calculated as:
    $ 
    \frac{0.588 - 0.523}{0.523} \times 100\% = 10.33\%.
    $
}
   \label{tab:regression_compare}
\end{table*}

\begin{table*}[b]
\centering
\begin{tabular}{l|ccccccccc}
\hline
Datasets     & Bagging & SSE  & sGBM & ANCL & NCL  & NCL* & SEA  & Improvement  \\
\hline
sonar        & 79.7    & 80.6 & 84.6 & 86.2 & \underline{86.7} & 86.2 & \textbf{88.2} & 1.73\% \\
ionosphere   & 88.4    & 87.4 & 91.9 & 93.5 & \underline{93.7} & 93.6 & \textbf{95.7} & 2.13\% \\
vehicle      & 81.1    & 80.4 & 82.8 & \underline{83.3} & 83.1 & 81.2 & \textbf{85.4} & 2.52\% \\
german       & 79.8    & 79.1 & 79.9 & 80.2 & \underline{80.3} & 80.1 & \textbf{82.6} & 2.86\% \\
dna          & 95.0    & 94.3 & 95.9 & \underline{96.0} & 95.9 & 95.9 & \textbf{97.8} & 1.88\% \\
satimage     & 90.8    & 89.4 & 90.7 & 88.1 & 90.7 & \underline{90.7} & \textbf{92.5} & 1.87\% \\
\hline
\end{tabular}
\caption{
    ACC results for classification tasks with ensemble sizes \(M \in \{5, 10, 20\}\), using 5-fold cross-validation.
    ACC values are presented without the percentage sign (\%).
    Bold values indicate the best ACC for each dataset, while underlined values represent the second-best ACC.
    ``Improvement" refers to the percentage improvement of the best ACC over the second-best ACC.
    For example, if the ACC improves from 86.7\% to 88.2\%, the relative improvement is calculated as:
    \(
    \frac{0.882-0.867}{0.867} \times 100\% = 1.73\%.
    \)
}
   \label{tab:classification_compare}
\end{table*}

\section{Experiments}\label{sec:experiments}

\subsection{Experimental Setup}\label{sec:ExperimentFramework}
This section outlines the datasets, evaluation metrics, and ensemble algorithms used for comparison.

\textbf{Datasets.}
We conducted comparative experiments on both regression and classification tasks, using 9 regression datasets (Table~\ref{tab:regression_datasets}) and 6 classification datasets (Table~\ref{tab:classification_datasets}). All datasets are publicly available and can be downloaded from LIBSVM \cite{CC01a}.
These datasets span various domains, such as healthcare, chemistry, and remote sensing. For example, the \textbf{pyrim}, \textbf{bodyfat}, and \textbf{dna} datasets are from the healthcare field. \textbf{Pyrim} contains chemical information related to drug metabolism, \textbf{bodyfat} predicts body fat percentage based on various body measurements, and \textbf{dna} records DNA splice site sequences used to classify DNA sequences into acceptor sites, donor sites, or non-splice sites based on 180 features.
From the chemistry domain, we used \textbf{triazines} and \textbf{mg}. The \textbf{triazines} dataset records the relationship between the chemical structure of triazine pesticides and their toxicity, while \textbf{mg} contains data related to chemical reactions involving magnesium.
For image processing and remote sensing, we used the \textbf{satimage} dataset, which consists of multispectral satellite imagery data. The task is to classify pixels into six land cover categories based on 36 features.
Additionally, we used datasets from other fields, such as the \textbf{cpusmall} dataset, which provides computer hardware performance data, and the \textbf{sonar} and \textbf{ionosphere} datasets, which involve signal propagation analysis.
Further details on all datasets are provided in Tables~\ref{tab:regression_datasets} and \ref{tab:classification_datasets}, and will not be elaborated here.
Additionally, these datasets exhibit diverse statistical properties, such as varying numbers of samples, features, and target classes, providing a comprehensive basis for evaluating the performance of different ensemble learning methods. Detailed information can be found in the tables.

\textbf{Evaluation Metrics.}    
We use Root Mean Squared Error (RMSE) as the evaluation metric for regression tasks and Accuracy (ACC) for classification tasks. RMSE and ACC are widely adopted metrics in their respective fields. RMSE measures the standard deviation between predicted values and true target values, and is defined as:
\begin{equation*}
    \text{RMSE}=\sqrt{
            \frac{
                \sum_{n=1}^{N}
                    (\hat{\bm{y}}^{(n)}-
                        t^{(n)})^2
             }{N}
    }.
\end{equation*}
\(N\) represents the total number of samples, \(\hat{\bm{y}}^{(n)}\) denotes the predicted value for the \(n\)-th sample, and \(t^{(n)}\) is the corresponding target value.

For classification tasks, ACC is defined as the proportion of correct predictions out of the total number of predictions. The formula for ACC is:
\begin{equation*}
    \text{ACC}=\frac{1}{N}\sum_{n=1}^{N}I(\hat{\bm{y}}^{(n)}=\bm{t}^{(n)}).
\end{equation*}
\(I(\cdot)\) is the indicator function, which returns 1 when the predicted value matches the true label, and \(\bm{t}^{(n)}\) represents the correct classification result for the \(n\)-th sample.

\textbf{Comparison Algorithms.}
We compare several classical ensemble algorithms as baselines. These methods are briefly described as follows: 
Bagging \cite{breiman1996bagging} generates multiple training subsets through bootstrap sampling, with each learner independently trained on these subsets. 
Soft Gradient Boosting Machine (sGBM) \cite{feng2020soft} modifies traditional GBM by parallelizing the training of learners instead of sequentially fitting residuals. This allows sGBM to optimize both local and global objectives simultaneously and has been shown to outperform classical ensemble methods like GBM and XGBoost. 
Snapshot Ensembles (SSE) \cite{huang2017snapshot} combines multiple learners found at different local minima using cyclical cosine annealing, which improves performance by capturing diverse local minima. 
NCL \cite{liu1997negatively,liu1999ensemble,liu1999simultaneous} balances individual accuracy and diversity explicitly by adjusting a penalty coefficient. 
NCL$^*$ \cite{brown2003use,brown2005managing} corrects a theoretical error in NCL and establishes the connection between NCL and ensemble error decomposition theory. 
AdaBoost Negative Correlation Learning (ANCL) \cite{wang2010negative} combines the strengths of NCL and AdaBoost by adjusting sample weights based on individual performance and diversity, significantly improving NCL's performance. 
Except for ANCL, which is specifically designed for classification tasks, all the above ensemble methods can be applied to both regression and classification tasks.

For classification tasks, we transform them into regression tasks using one-hot encoding \cite{bishop2006pattern,hastie2009elements}. For a classification variable with \( K \) categories, one-hot encoding represents each category as a binary vector of length \( K \), where the position corresponding to the category is set to 1, and all other positions are set to 0. This enables regression models to predict these binary vectors, allowing classification problems to be addressed using regression models.

\begin{figure}[tb]
    \centering
    \includegraphics[width=0.486\textwidth]{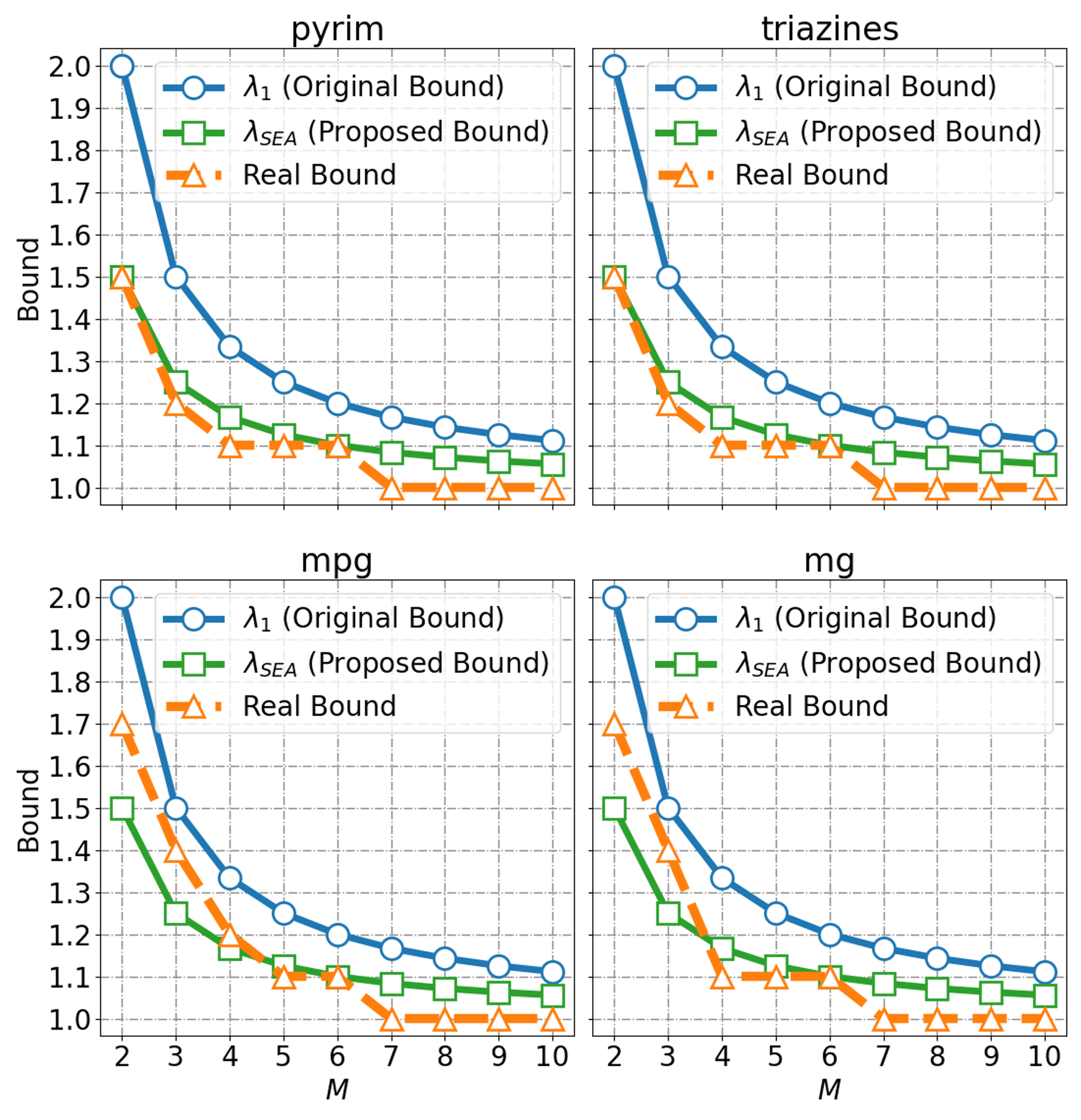}
    \caption{
        Comparison of NCL's original theoretical boundary ($\lambda_1$, original bound), the proposed theoretical boundary based on the SEA framework ($\lambda_{SEA}$, proposed bound), and the real boundary (real bound). 
        We randomly selected four datasets to compare the theoretical and real boundaries. 
        Note that the adjustable parameter is incremented by 0.1, providing an approximate estimate of the theoretical boundary, which causes the real boundary to exhibit a step-like structure.
    }
    \label{fig:realTheoryBound}
\end{figure}

\textbf{Algorithm Setup.}
For each dataset, all ensemble methods use the same individual learners and random seed. For regression tasks, we use a multilayer perceptron (MLP) as the base model, with sigmoid activation and 3 layers. For classification tasks, we use either MLP or LeNet5 \cite{lecun1998gradient} as the base model, with the same MLP configuration as mentioned above. 
All experiments were conducted on a server running Ubuntu 20.04.4 LTS, equipped with an AMD EPYC 9554 processor and an NVIDIA A800 GPU.

The training setup for the ensemble methods follows the configurations specified in the original papers and relevant literature. For Snapshot Ensembles, snapshots were taken every 60 epochs. For NCL, NCL$^*$, ANCL, and SEA, the corresponding adjustable parameters were set as follows:
NCL: \(\lambda \in \{0, 0.1, \ldots, 1\}\), 
NCL$^*$: \(\gamma \in \{0, 0.1, \ldots, 1\}\), 
ANCL: \(\eta \in \{0.25, 1.2, \ldots, 12\}\), 
SEA: \({k \in \{0, 0.1, \ldots, 2\}}\).

\subsection{Performance Analysis}\label{sec:PerformanceAnalysis}

The experiments were conducted using 5-fold cross-validation with an 80\% training and 20\% testing data split. We studied the impact of ensemble size, with \(M \in \{5, 10, 20\}\), on different ensemble methods. The reported results represent the mean performance across 15 experiments (5 folds \(\times\) 3 ensemble sizes).
The results for regression and classification tasks are shown in Table~\ref{tab:regression_compare} and Table~\ref{tab:classification_compare}, respectively. The following conclusions can be drawn:

Overall, adjustable ensemble methods (NCL, NCL$^*$, ANCL, SEA) outperform general ensemble methods (Bagging, SSE, sGBM), highlighting the superiority of adjustable ensemble methods.
Among these, SEA performs the best. For most regression datasets, SEA's RMSE improves by 10\% to 20\% compared to the best-performing baseline, while on classification datasets, SEA's ACC shows a 2\% to 3\% improvement over the best baseline, demonstrating SEA's effectiveness.
Notably, across most datasets, the performance trend follows 
$SEA \succ NCL \succ NCL^*$, 
which is consistent with the theoretical findings discussed earlier: differences in their effective adjustment ranges lead to corresponding performance differences (Figure~\ref{fig:range}).

\begin{figure*}[b]
    \centering
    \includegraphics[width=0.70\textwidth]{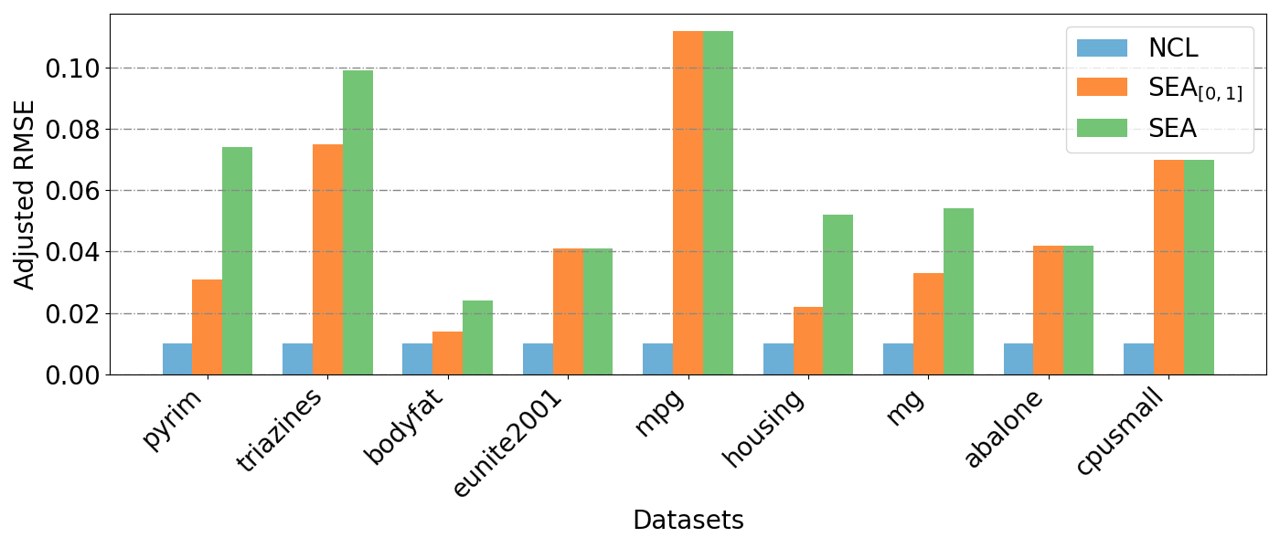}
    \caption{
        Comparison of the performance of NCL, SEA$_{[0,1]}$, and SEA.
        For clarity, we introduce Adjusted RMSE, where a larger value indicates better performance relative to the other two models.
        The formula for Adjusted RMSE is:
        \(
        \text{RMSE}_{adjust} = \text{RMSE}_{max} - \text{RMSE} + 0.01,
        \)
        designed to more clearly highlight the performance differences between the three models.
    }
    \label{fig:NCL-SEA01-SEA}
\end{figure*}

\begin{figure*}[b]
    \centering
    \includegraphics[width=0.80\textwidth]{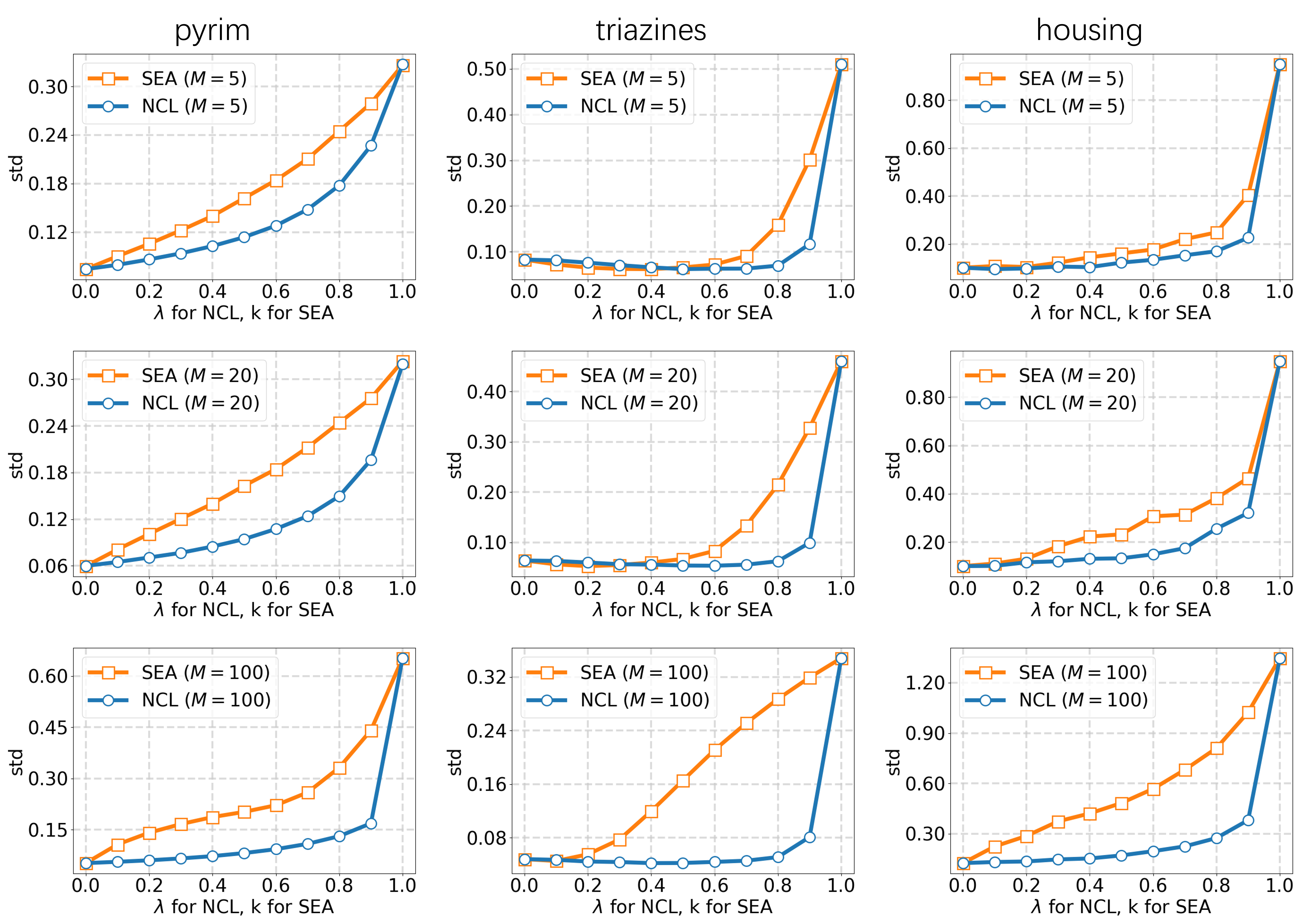}
    \caption{
        The standard deviation (\(\text{std}\)) of NCL and SEA as a function of the adjustable parameter (\(\lambda\) or \(k\)) on three randomly selected regression datasets, with ensemble sizes \(M \in \{5, 20, 100\}\). 
    }
    \label{fig:uni-std}
\end{figure*}

\subsection{Theoretical Boundary Validation}
\label{sec:definition}

In previous research, we redefined the boundary conditions for adjustable ensemble methods: 
(1) The loss function must be optimizable; 
(2) The ensemble error must decrease after training.
Traditional methods typically only consider the first boundary condition, but by incorporating both conditions, we can derive tighter boundaries. 

Through the SEA framework, we can efficiently calculate the boundaries for general adjustable ensemble methods. For instance, in the classical adjustable ensemble method NCL, when considering only boundary condition (1)—i.e., using the positive definiteness of the Hessian matrix—the original NCL boundary is:
\[
\lambda_1 < \frac{M}{M-1}.
\]
However, the SEA framework yields a tighter boundary:
\[
\lambda_{SEA} < \frac{2M - 1}{2(M - 1)}.
\]
Figure~\ref{fig:realTheoryBound} illustrates the relationship between the original theoretical boundary ($\lambda_1$), the proposed theoretical boundary ($\lambda_{SEA}$), and the real boundary as the ensemble size $M$ increases.

The criteria for determining the real boundary are as follows: as the adjustable parameter increases, performance eventually converges to a lower level and stabilizes. We define this stabilized range as the ``low-performance region". The boundary point is defined as the critical point just before entering the low-performance region. Numerically, we set the critical point as having at least a 5\% improvement in performance compared to the low-performance region. 
As shown in Figure~\ref{fig:realTheoryBound}, the real boundary is closer to our proposed boundary than to the original boundary.

\subsection{Ablation Study}\label{sec:advantage}

In this section, we investigate the advantages of SEA’s adjustment strategy through ablation experiments, going beyond its broader adjustment range. To ensure a fair comparison, we constrain SEA’s adjustment range to match that of NCL. Specifically, when $k \in [0,1]$, both methods have the same effective adjustment range. This equivalence can be derived from the relationship between SEA and NCL’s adjustable parameters (see Eq.~\ref{eq:k_NCLorginal}). The version of SEA with this limited range is denoted as SEA$_{[0,1]}$.

Figure~\ref{fig:NCL-SEA01-SEA} presents a performance comparison between NCL, SEA$_{[0,1]}$, and the original SEA across multiple datasets. The overall results can be summarized as follows:
\[
SEA \succ SEA_{[0,1]} \succ NCL.
\]
The superiority of
\(
SEA \succ SEA_{[0,1]}
\)
is primarily attributed to SEA’s larger effective adjustment range:
\(
R_{SEA}=[0,2] \supset R_{SEA_{[0,1]}}=[0,1].
\)

In the case of
\(
SEA_{[0,1]} \succ NCL,
\)
despite having the same adjustment range ($R_{SEA_{[0,1]}} = R_{NCL}$), SEA$_{[0,1]}$ consistently outperforms NCL. We hypothesize that this is due to SEA$_{[0,1]}$ employing a more effective adjustment strategy: SEA introduces more "uniform" changes in diversity relative to the adjustable parameters compared to NCL. This uniformity likely enables SEA to explore a broader range of model performance and diversity combinations, increasing the chances of finding an optimal ensemble configuration.

In regression tasks, diversity is typically measured by the standard deviation (\(\text{std}\)) among individual learners. To further analyze diversity, we studied the relationship between the adjustable parameters and the corresponding \(\text{std}\) for both NCL and SEA. Figure~\ref{fig:uni-std} shows how the \(\text{std}\) changes as the adjustable parameters vary. Across all datasets and ensemble sizes, SEA displays more uniform changes in \(\text{std}\) compared to NCL, indicating a more stable diversity adjustment. This effect becomes even more pronounced as the ensemble size increases.

This observation aligns with the theoretical derivation in Section~\ref{app:theorem}, where SEA's relationship between \(\text{std}\) and the adjustable parameters is approximately linear, whereas NCL exhibits a nonlinear relationship. Moreover, NCL's curvature becomes more significant as the ensemble size $M$ increases. For instance, in the triazines dataset, the curvature of $\text{std}(\lambda)|_{M=100}$ is considerably larger than that of $\text{std}(\lambda)|_{M=5}$.

\section{Conclusion and Future Work}
\label{sec:conclusion}

This paper introduces a novel Self-Error Adjustment (SEA) framework that allows for precise control of the balance between individual learner performance and diversity in ensemble learning. By decomposing ensemble error from a training perspective and incorporating adjustable parameters into the loss function, SEA facilitates flexible adjustment of ensemble performance. 
Compared to traditional Negative Correlation Learning (NCL) and its variants, SEA provides a wider effective adjustment range and achieves more uniform diversity changes. Theoretically, we derived tighter boundaries for general adjustable ensemble methods, demonstrating SEA's advantages in terms of effective adjustment range and diversity modulation. Experimental results across multiple public regression and classification datasets validate SEA's superior performance over baseline methods. Ablation studies further confirmed that SEA's enhanced performance is driven by its broader adjustment range and smoother diversity changes.
Despite the significant performance gains SEA achieves across various tasks, several avenues remain for future research. First, validating SEA’s performance on larger datasets and more complex models. Second, exploring the application of the SEA framework to different types of ensemble learning models, such as ensembles of large language models. Lastly, investigating methods to automatically select or adjust SEA’s parameters to better suit varying data and task requirements.

%%%%%%%%% BODY TEXT %%%%%%%%%%%%%%%%%%%%%%%%%%%%%%%%%%%%%%%%

% \cleardoublepage
% \clearpage

\bibliographystyle{IEEEtran} % elsarticle-harv是目标期刊的参考文献格式
\bibliography{references}  % references是bib文件的文件名

\end{document}